\pdfoutput=1
\documentclass[runningheads]{llncs}
\usepackage[T1]{fontenc}
\usepackage{graphicx}
\usepackage{amsmath}
\usepackage{amssymb} 
\usepackage{booktabs}
\usepackage{multirow}
\usepackage{graphicx}
\usepackage{optidef}
\usepackage[ruled,vlined,linesnumbered]{algorithm2e}
\usepackage{subcaption} 

\begin{document}
\title{Lateral Reciprocal Collision Avoidance: A Probabilistic Social-Norm-Inspired Strategy for Deadlock-Free Multi-Robot Navigation}
%
%
\author{Siyi Lu\inst{1}\orcidID{0000-0003-4637-844X} \and
Sipu Ruan\inst{*1}\orcidID{0000-0001-5075-6728}}
\authorrunning{F. Author et al.}
%
\institute{Beihang University, Beijing, China \\
\email{\{siyilu,ruansp\}@buaa.edu.cn}}
\maketitle              
\begin{abstract}
Safe and efficient obstacle avoidance in multi-robot navigation is a challenging problem, with deadlock being a key issue due to strict collision-free constraints. This paper proposes a novel Lateral Reciprocal Collision Avoidance (LRCA) strategy based on velocity obstacle theory to mitigate deadlock among multiple robots. Inspired by pedestrian collision avoidance, LRCA incorporates symmetric lateral displacement to resolve deadlock. Unlike Optimal Reciprocal Collision Avoidance (ORCA) algorithm, which computes velocity constraints based on the minimal adjustment across the full velocity obstacle, our proposed method restricts the velocity changes of the agents to one randomly selected side of the relative velocity. This randomized directional selection strategy effectively prevents deadlock while preserving collision avoidance. This approach avoids conflicting velocity changes that could lead to mutual trapping. Theoretical analysis shows how ORCA cause deadlock using quadratic programming, Lagrangian functions, and KKT conditions, and how LRCA effectively prevents this. Extensive simulations across four benchmark multi-robot navigation scenarios show that LRCA outperforms existing algorithms in success rate, time to goal, path length, and computational efficiency.

\end{abstract}
\section{Introduction}
Safe and efficient multi-robot navigation is a fundamental problem in robotics and plays a critical role in a wide range of service and industrial applications \cite{Wang05052025}. In these task, multiple robots operate in a shared environment and are required to reach their respective goals while avoiding collisions with each other robots \cite{b2}. Designing navigation strategies that guarantee collision-free behavior while maintaining efficiency remains a challenging problem.
Existing approaches to multi-robot navigation can be broadly categorized into centralized and decentralized methods. Centralized methods compute control actions for all robots jointly, which allows them to provide strong guarantees on safety, completeness, and near-optimality \cite{b3,b4,b5}. However, these methods typically suffer from high computational costs and rely on reliable, synchronized communication among all robots. Consequently, they are difficult to scale to large systems and are vulnerable to single-point failures. In contrast, decentralized methods allow each robot to independently determine its control actions based only on locally observable information, such as the positions, velocities, and shapes of neighboring robots \cite{b6,b7,b8,b9,b10}. These methods are computationally lightweight, more robust to failures, and exhibit better generalization across different environments, making them more suitable for large-scale and real-world multi-robot systems.

\begin{figure}[!t]
    \centering
    \includegraphics[width=1\linewidth]{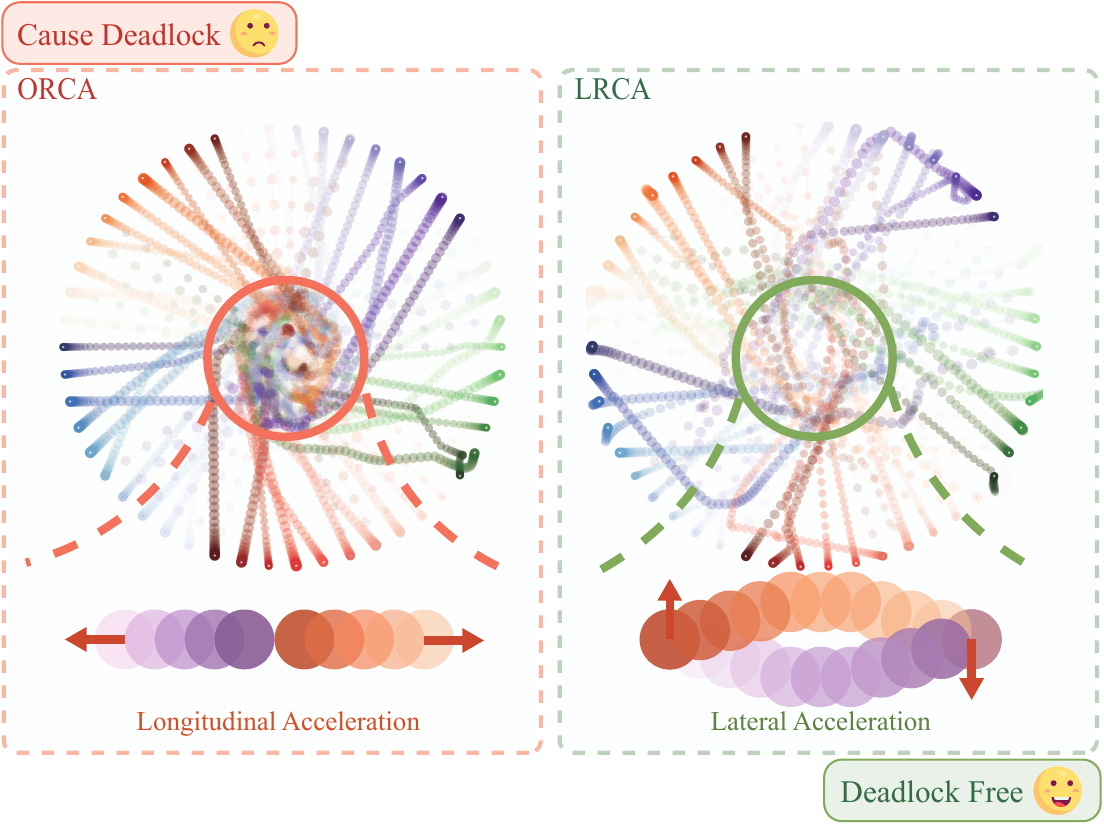}
    \caption{The left side shows 50 robots navigating with the ORCA, while the right side shows 50 robots navigating with the LRCA. The LRCA method utilizes lateral acceleration, rather than relying solely on longitudinal acceleration as in the ORCA method, which is the primary cause of deadlock.}
    \label{fig:moti}
\end{figure}

Among decentralized approaches, the Velocity Obstacle (VO) \cite{b11} framework has become a widely used tool for collision detection, as it enables efficient prediction of potential collisions between pairs of robots within a given time horizon. Building on the VO framework, Reciprocal Collision Avoidance (RCA) \cite{b12} methods have been proposed to share collision avoidance responsibility among robots, resulting in efficient and scalable navigation strategies. Despite their practical success, RCA-based methods suffer from a critical limitation: deadlock \cite{b13}. Deadlock significantly degrades navigation success rates and efficiency and may occur even in simple scenarios involving only two robots, posing a serious challenge for decentralized multi-robot navigation.
From an optimization perspective, RCA methods can be interpreted as VO-based quadratic programs(QPs). In multi-robot navigation, two objectives naturally coexist: reaching the goal and avoiding collisions. In VO-based QPs, collision avoidance is typically enforced as a hard constraint, while progress towards the goal is treated as the optimization objective. By prioritizing safety over task completion, this formulation guarantees collision-free behavior but may result in deadlock, in which robots remain safe yet make no progress toward their goals. In this paper, we analyze the emergence of deadlock in RCA frameworks from the perspective of VO-based QPs and provide a theoretical characterization of deadlock scenarios using the Karush–Kuhn–Tucker (KKT) conditions.

In contrast to robotic systems, human crowds exhibit remarkably robust collision avoidance behavior and rarely experience deadlock, even in dense environments \cite{b14}. Pedestrian avoidance behavior is inherently decentralized, as individuals make decisions based on local observations of each other and shared conventions, which is owe commonly referred to as "social norms". These social norms correspond to an implicit assumption in decentralized robot navigation: \textit{all agents are aware that others follow the same avoidance strategy}. One particularly effective social norm observed in pedestrian dynamics is symmetric lateral displacement, where individuals consistently sidestep in a coordinated manner to avoid collisions \cite{b15}. This shared convention plays a crucial role for human crowds to prevent deadlock.
Motivated by this observation, we propose to incorporate pedestrian social norms into the RCA framework to mitigate deadlock effects in multi-robot navigation. Specifically, as shown in Fig. \ref{fig:moti}, we introduce ``Lateral Reciprocal Collision Avoidance (LRCA)'', a novel RCA-based strategy that integrates symmetric lateral displacement as a shared consensus among robots. By embedding this social norm into the VO-based decision-making process, LRCA significantly reduces the occurrence of deadlock while preserving the decentralized nature of the navigation strategy. Furthermore, we provide a theoretical analysis based on VO-based QPs and KKT conditions to show that LRCA effectively avoids deadlock.
The main contributions of this paper are as follows. 

\begin{itemize}
    \item We formalize the RCA framework as a VO-based QPs and provide a theoretical analysis of deadlock scenarios using KKT conditions.
    \item We propose LRCA, a novel RCA framework inspired by pedestrian lateral avoidance behavior.
    \item We prove and validate LRCA's ability to avoid deadlock in several scenarios prone to ORCA deadlock. 
    
\end{itemize}

\section{Literature Review}
This section reviews prior work related to multi-robot collision avoidance. 
We first summarize centralized and decentralized collision avoidance methods, with particular emphasis on cooperative decentralized approaches, and then review research on social norms in decentralized collision avoidance.

\subsection{Centralized and Decentralized Collision Avoidance}

Collision avoidance is essential for collision-free motion planning in multi-robot systems. Approaches are typically categorized as centralized and decentralized. Centralized methods, such as Conflict-Based Search (CBS) \cite{b17}, plan trajectories using complete system information, including unobservable attributes like goal locations \cite{b16}. Variants like explicit estimation CBS (EECBS) \cite{b18}, lazy constraints addition search
for MAPF (LaCAM) \cite{b19}, and priority inheritance with backtracking (PIBT) \cite{b20} improve efficiency and scalability but rely on synchronized communication, limiting scalability in large systems.
Decentralized methods use only local states for control, without robot communication. These methods are further divided into non-cooperative and cooperative approaches. Non-cooperative methods, like SARL \cite{b23} and T-MPC \cite{b24,b25}, rely on behavior prediction, while cooperative methods assume shared collision avoidance strategies, improving efficiency and safety \cite{b26,b27}.

Among cooperative methods, velocity obstacle (VO)-based approaches \cite{b11} are widely adopted. The Reciprocal Velocity Obstacle (RVO) \cite{b28} extends VO by assuming reciprocal responsibility for collision avoidance, with ORCA \cite{b10} defining admissible velocity spaces via linear constraints. While VO-based methods are computationally efficient, they can lead to deadlock in certain scenarios, highlighting the need for further theoretical investigation \cite{b13}. Our proposed LRCA algorithm can be categorized as a cooperative method and specifically mitigates the deadlock phenomena inherent in RVO-based methods. While retaining the computational efficiency of VO-based approaches, it achieves safety and navigation performance comparable to, or even exceeding, learning-based methods.

\subsection{Social Norms in Decentralized Collision Avoidance}
A crowd consists of many individuals coexisting in the same space and time, whose motions are primarily governed by local interactions over extended periods. Pedestrians typically perform operational movements toward tentative destinations that may change over time and are not bound to arrive at specific locations at fixed times \cite{b29}. Empirical studies show that pedestrian collision avoidance emerges through early trajectory adjustments made several meters before potential conflict points, reducing interactions and preventing collisions. Since pedestrians do not have access to each other’s destinations, crowd navigation can be viewed as a decentralized multi-agent collision avoidance problem.

Despite the absence of centralized coordination, human crowds achieve efficient collision avoidance through a hierarchical structure in which shared social norms guide tactical side selection and operational control handles speed adjustment \cite{b30}. In game-theoretic terms, a convention corresponds to one of multiple stable equilibria in a game with two or more equilibria \cite{b31}, and such conventions are prevalent in pedestrian dynamics, for example, the tendency to walk on a preferred side \cite{b15}. These shared conventions enable crowds to function as a cooperative decentralized navigation system. As an intelligent system, human crowds provide valuable insights for multi-robot navigation. Motivated by the widely observed “keep-right” rule in pedestrian behavior, this paper leverages lateral social conventions to propose a novel lateral reciprocal collision avoidance strategy that addresses deadlock in multi-robot navigation.

\section{Problem Definition}
In a multi-robot navigation scenario with $N$ robots, each robot is disc-shaped moving in a two-dimensional plane $\mathbb{R}^2$. The robots' attributes are divided into external and internal properties. External properties $\mathcal{O}_e=\{\boldsymbol{p},\boldsymbol{v},r\}$ include position $\boldsymbol{p}$, velocity $\boldsymbol{v}$, disc-shape, and radius $r$, while internal properties $\mathcal{O}_i=\{\boldsymbol{p}^{goal},\boldsymbol{v}^{pref}\}$ include the goal location $\boldsymbol{p}^{goal}$ and preferred velocity $\boldsymbol{v}^{pref}$ (the robot's speed in the absence of other influential factors).
At each time step $\tau$, each robot independently selects a new velocity $\boldsymbol{v}^{new}$ based on the external attributes of other robots, assuming that all robots follow the same cooperative collision avoidance strategy. The robot adjusts its velocity to avoid collisions while steering towards its target location, moving with its preferred velocity.

The velocity obstacle $VO_{i|j}^\tau$ is a set of relative velocities $\boldsymbol{v}_{i|j}$ between two robots, $i$ and $j$, that would result in a collision within the time interval $[0,\tau]$. Let $D(\boldsymbol{p}, r)$ denote an open disc of radius $r$ centered at $\boldsymbol{p}$, i.e.,
\begin{equation}
D(\boldsymbol{p}, r)\doteq\{\boldsymbol{q} \mid\|\boldsymbol{q}-\boldsymbol{p}\|<r\},
\end{equation}
then:
\begin{equation}
V O_{i \mid j}^\tau=\left\{\boldsymbol{v}_{i|j} \mid \exists t \in[0, \tau]:: t \boldsymbol{v}_{i|j} \in D\left(\boldsymbol{p}_j-\boldsymbol{p}_i, r_i+r_j\right)\right\}.
\end{equation}

In cooperative reciprocal collision avoidance \cite{b10}, the set of permitted velocities $V_i$ for robot $i$ and $V_j$ for robots $j$ are determined based on their velocity obstacles $V O_{i \mid j}^\tau$:
\begin{equation}
V_i \cap (V O_{i \mid j}^\tau \oplus V_j)=\emptyset,
\end{equation}
where $\oplus$ represents Minkowski sum. The union of the admissible velocity sets for robot $i$ with respect to all other robots defines the set of velocities that allow robot $i$ to avoid collisions in the next time step.

\section{Lateral Reciprocal Collision Avoidance}
\label{sec:4}
In the proposed framework, we aim to define sets of permitted velocities $V_i$ for robot $i$ and $V_j$ for robot $j$, such that these sets ensure reciprocal collision avoidance within a given time step $\tau$. The goal is to maximize the proximity of the admissible velocities to the robots' optimal velocities, while maintaining collision-free motion.
To determine the sets of permitted velocities for both robots, we first calculate their relative velocity $\boldsymbol{v}_{i\mid j}$. Then we compute the smallest adjustment vector $\boldsymbol{u}$ needed to avoid or prevent a collision. This adjustment vector is determined as the vector from the relative velocity $\boldsymbol{v}_{i\mid j}$ difference to the one point on the boundary of the velocity obstacle. The classic ORCA \cite{b10} selects the point on the velocity obstacle $VO_{i|j}^\tau$ boundary that is closest in terms of relative velocity $\boldsymbol{v}_{i\mid j}$, but it can lead to severe deadlocks. 

\subsection{Deadlock-Free Collision Avoidance}
Inspired by pedestrian collision avoidance behavior, we constrain the relative velocity adjustment $\delta\boldsymbol{v}$ to avoid being parallel to the current relative velocity, thereby preventing deadlock.
Let the updated velocities of robots $i$ and $j$ be
\begin{equation}
\boldsymbol{v}_i^{\mathrm{new}} = \boldsymbol{v}_i^{\mathrm{opt}} + \delta\boldsymbol{v}_i, \qquad
\boldsymbol{v}_j^{\mathrm{new}} = \boldsymbol{v}_j^{\mathrm{opt}} + \delta\boldsymbol{v}_j .
\end{equation}

The new relative velocity on next timestep is given by
\begin{equation}
\boldsymbol{v}_{i|j}^{\mathrm{new}}
= \boldsymbol{v}_i^{\mathrm{new}} - \boldsymbol{v}_j^{\mathrm{new}}
= \boldsymbol{v}_{i|j}^{\mathrm{opt}} + \left( \delta\boldsymbol{v}_i - \delta\boldsymbol{v}_j \right).
\end{equation}

We impose the following lateral constraint on the relative velocity adjustment:
\begin{equation}
\forall \, k \in \mathbb{R}, \quad
\delta\boldsymbol{v}_i - \delta\boldsymbol{v}_j
\neq k \, \boldsymbol{v}_{i|j}^{\mathrm{opt}} .
\end{equation}

Under this constraint, the adjustment term cannot exactly cancel the current relative velocity. Therefore, when $\boldsymbol{v}_{i|j}^{\mathrm{opt}} \neq \boldsymbol{0}$, the updated relative velocity satisfies
\begin{equation}
\boldsymbol{v}_{i|j}^{\mathrm{new}} \neq \boldsymbol{0}.
\end{equation}

\begin{figure}[!t]
    \centering
    \includegraphics[width=1\linewidth]{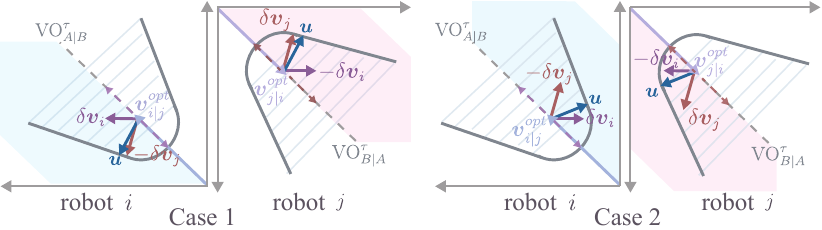}
    \caption{Two cases ensuring the velocity adjustment difference is not parallel to the relative velocity: Case 1: Both robots' adjustments are counterclockwise; Case 2: Both are clockwise.}
    \label{fig:two_ways}
\end{figure}

A more rigorous and complete derivation process is in Sec. \ref{sec:prov_lrca}. This lateral adjustment avoids collision resolution that relies solely on accelerating or decelerating along the relative velocity direction, which is prone to inducing deadlock. 
To prevent the difference in velocity adjustments between two robots $\delta\boldsymbol{v}_i-\delta\boldsymbol{v}_j$ from being parallel to the relative velocity $\boldsymbol{v}_i^{opt}-\boldsymbol{v}_j^{opt}$, two cases are considered: both robots' velocity adjustments can be in the clockwise direction of the relative velocity, or both can be in the counterclockwise direction, as shown in Fig. \ref{fig:two_ways}. Note that for robot $j$, the relative velocity with respect to robot $i$ is calculated as $\boldsymbol{v}_j^{opt}-\boldsymbol{v}_i^{opt}$.

Since LRCA introduces constraints on the velocity adjustments, requiring them to be either in the clockwise or counterclockwise direction of the relative velocity, continuing to select the minimum velocity adjustment to avoid collisions would shrink the feasible velocity set, potentially even resulting in an empty set with no valid velocities. This could lead to another form of deadlock. To address this, we select the velocity adjustment with the largest angle relative to the relative velocity, which is the velocity perpendicular to the linear boundary of the $V O_{A \mid B}^\tau$, thereby maximizing the feasible velocity set while ensuring collision avoidance. As shown in Fig. \ref{fig:two_ways}, the velocity adjustment can either be perpendicular to the left linear boundary of the $V O_{i \mid j}^\tau$ or perpendicular to the right linear boundary, corresponding to two strategies for avoiding deadlock. Let $\boldsymbol{n}$ be the outward normal of the boundary of $V O_{i \mid j}^\tau$ at point $\boldsymbol{v}_{i|j}^{opt}$ + $\boldsymbol{u}$, the relative velocity adjustment $\boldsymbol{u}$ can be formally expressed as:

\begin{equation}
    \boldsymbol{u} = \left( \boldsymbol{n}^\top \left( \boldsymbol{v}_i^{\mathrm{opt}} - \boldsymbol{v}_j^{\mathrm{opt}} \right) \right) \boldsymbol{n},
\end{equation}
where 
\begin{equation}
\label{equa:min_u}
\begin{aligned}
 &\quad \boldsymbol{n} = 
\begin{cases}
    \mathbf{R}\left( \frac{\pi}{2} + \theta \right) \boldsymbol{e}_{ij}, & \text{(Case 1)}, \\
    \mathbf{R}\left( -\frac{\pi}{2} - \theta \right) \boldsymbol{e}_{ij}, & \text{(Case 2)},
\end{cases} \\
&\quad \theta = \arcsin\left( \frac{r_i + r_j}{\|\boldsymbol{p}_j - \boldsymbol{p}_i\|_2} \right), \\
 &\quad \boldsymbol{e}_{ij} = \dfrac{\boldsymbol{p}_j - \boldsymbol{p}_i}{\|\boldsymbol{p}_j - \boldsymbol{p}_i\|_2}, \\
\end{aligned}
\end{equation}
and $\mathbf{R}\left( \frac{\pi}{2} + \theta \right)$ and $\mathbf{R}\left( -\frac{\pi}{2} - \theta \right)$ are rotation matrices that rotates $\boldsymbol{e}_{ij}$ counterclockwise and clockwise by $\frac{\pi}{2} + \theta$ degrees respectively. Then, the set $\mathrm{LRCA}_{i \mid j}^\tau$ of permitted velocities for robot $i$ can be formally expressed as:

\begin{equation}
\label{eq:lrca}
\mathrm{LRCA}_{i \mid j}^\tau(k) = 
\left\{ \boldsymbol{v} \;\middle|\; 
\begin{aligned}
    &\left( \boldsymbol{v} - \left( \boldsymbol{v}_i^{\mathrm{opt}} + \frac{1}{2} \boldsymbol{u} \right) \right) \cdot \boldsymbol{n} \geq 0, \\
    &\mathrm{sign}\left( \left( \boldsymbol{v} - \boldsymbol{v}_i^{\mathrm{opt}} \right) \times \boldsymbol{v}_{i|j}^{\mathrm{opt}} \right) = s_k
\end{aligned} \right\},
\quad k=1,2,
\end{equation}
where $s_1 = 1$ (corresponding to Case 1), $s_2 = -1$ (corresponding to Case 2). The specific case is chosen by the randomized strategy in Sec. \ref{sec:random}.

\subsection{Game-Theoretic Randomization for Distributed Collision Avoidance}
\label{sec:random}
To enable robots to reach consensus without explicit communication, selecting either Case 1 or Case 2 as shown in Fig. \ref{fig:two_ways}, we introduce a randomized mechanism inspired by human game-theoretic interactions. Each robot $i$ randomly selects whether the velocity adjustment $\delta\boldsymbol{v}_i$ lies on the clockwise or counterclockwise side of the relative velocity with respect to robot $j$. If both robots $i$ and robot $j$ choose the same side, collision avoidance is achieved without deadlock; otherwise, avoidance is not possible. Alg. \ref{alg:LRCA} outlines an implicit coordination process that enables the robots to reach consensus without communication.

\begin{algorithm}
\caption{Algorithm process of LRCA}
\label{alg:LRCA}
\SetAlgoLined
\KwIn{Number of robots $n$, positions $\mathbb{P}$, velocities $\mathbb{V}^{\mathrm{opt}}$, attributes $\mathbb{A}$}
\KwOut{New velocities $\mathbb{V}^{\mathrm{new}}$ for next time step}
\For{$i = 1$ \KwTo $n$}{
    \For{$j = 1$ \KwTo $n, j \neq i$}{
        Compute velocity obstacle $VO_{i|j}^\tau$ and relative velocity $\boldsymbol{v}_{i|j}^{\mathrm{opt}}$\;
        \If{$\boldsymbol{v}_{i \mid j}^{\mathrm{opt}} \in VO_{i|j}^\tau$}{
            Calculate velocity changes $\delta\boldsymbol{v}_i$ and $\delta\boldsymbol{v}_j$\;
            \eIf{$(\boldsymbol{v}_{i \mid j}^{\mathrm{opt}} \times \delta\boldsymbol{v}_i) \cdot (\boldsymbol{v}_{i \mid j}^{\mathrm{opt}} \times \delta\boldsymbol{v}_j) < 0$}{
                \eIf{$\boldsymbol{v}_{i \mid j}^{\mathrm{opt}} \times \delta\boldsymbol{v}_i > 0$}{
                    Add Case 1 constraint\;
                }{
                    Add Case 2 constraint\;
                }
            }{
                Randomly add Case 1 or Case 2 constraint\;
            }
        }
    }
}
\end{algorithm}

In Algorithm \ref{alg:LRCA}, lines 1–2 iterate over all robot pairs $(i, j)$ for distributed pairwise collision checking.
Line 3 constructs the velocity obstacle $VO_{i|j}^\tau$ and computes the nominal relative velocity $\boldsymbol{v}_{i|j}^{\mathrm{opt}}$.
Line 4 detects potential collisions by checking whether $\boldsymbol{v}_{i|j}^{\mathrm{opt}}$ lies inside $VO_{i|j}^\tau$.
Line 5 computes candidate velocity adjustments $\delta\boldsymbol{v}_i$ and $\delta\boldsymbol{v}_j$.
Line 6 determines whether the two robots selected the same lateral side using the sign of the cross product.
Lines 7–11 add the corresponding deterministic lateral constraint (Case 1 or Case 2) when a consistent side is detected.
Line 13 randomly selects between Case 1 and Case 2 according to Eq. \eqref{alg:probability} when the selections are inconsistent. Let the random variable $C$ denote the selected category, where $C=1$ corresponds to Case 1 and $C=2$ corresponds to Case 2. Define the angles:


\begin{equation}
\theta_1=\angle\left(\boldsymbol{n}_1,-\boldsymbol{v}_{i \mid j}^{o p t}\right), \quad \theta_2=\angle\left(\boldsymbol{n}_2,-\boldsymbol{v}_{i \mid j}^{o p t}\right),
\end{equation}
where $\boldsymbol{n}_1=\mathbf{R}\left( \frac{\pi}{2} + \theta \right) \boldsymbol{e}_{ij}$ and $\boldsymbol{n}_2=\mathbf{R}\left( -\frac{\pi}{2} - \theta \right) \boldsymbol{e}_{ij}$,  $\angle(\boldsymbol{a},\boldsymbol{b})$ represents the angle between vector $\boldsymbol{a}$ and vector $\boldsymbol{b}$. Then $C$ follows a Categorical distribution with parameters $p_1=\frac{\theta_2}{\theta_1+\theta_2}$ and $p_2=\frac{\theta_2}{\theta_1+\theta_2}$, i.e.:
\begin{equation}
\label{alg:probability}
    C \sim \text{Categorical}\left(p_1, p_2\right).
\end{equation}

As the relative velocity $\boldsymbol{v}_{i\mid j}^{opt}$ approaches the left boundary of the $VO_{i\mid j}^\tau$, $\theta_1$ and $p_1$ increase. Conversely, as the relative velocity $\boldsymbol{v}_{i\mid j}^{opt}$ approaches the right boundary of the $VO_{i\mid j}^\tau$, $\theta_2$ and $p_2$ increase.

\section{Analysis of the Two-Robot Deadlock}
In Sec.~\ref{sec:4}, we introduced the LRCA method from a set-theoretic perspective. In this section, we reformulate LRCA and ORCA using quadratic programming and present their respective Lagrangian functions and KKT conditions, applying them to a two-robot example. This analysis reveals the underlying characteristics of deadlock in ORCA and demonstrates how LRCA effectively prevents deadlock. These findings can be extended to scenarios involving $N$ robots. The velocity constraints for each robot in the ORCA and LRCA algorithms are given by Eq. \eqref{equa:min_u_orca} and the combination of Eq. \eqref{equa:min_u_orca} and Eq. \eqref{equa:rela_v}, respectively. 

\begin{equation}
\label{equa:min_u_orca}
\boldsymbol{n}^\top\boldsymbol{v}_i^{new}\ge\boldsymbol{n}^\top\boldsymbol{v}_i^{opt}+\frac{\left\|\boldsymbol{u}\right\|_2}{2},
\end{equation}

\begin{equation}
    \label{equa:rela_v}
        \boldsymbol{h}^\top\boldsymbol{v}_i^{new} \le \boldsymbol{h}^\top\boldsymbol{v}_i^{opt},
\end{equation}
where $\boldsymbol{h}=\angle(\boldsymbol{v}_{i\mid j}^{opt},\frac{\pi}{2})$ or $\boldsymbol{h}=\angle(\boldsymbol{v}_{i\mid j}^{opt},-\frac{\pi}{2})$ according to the randomized decision strategy described
in Sec. \ref{sec:random}. Therefore, any $\boldsymbol{v}_i^{new}$ generated by ORCA or LRCA that satisfy corresponding constraint are guaranteed to ensure collision-free and deadlock-free trajectories for robots $i$ and $j$ in the multi-robot system. Since robot $i$ wants to avoid collisions with $N-1$ robots, there are $N-1$ collision avoidance constraints. To mediate between safety and goal stabilization objective, a QP is posed that computes a controller closest to the target velocity $\hat{\boldsymbol{v}}_i\left(p_i\right)$ and satisfies the $N-1$ constraints, as shown:

\begin{equation}
\label{equa:qp}
\boxed{
\begin{aligned}
    \underset{\boldsymbol{v}_i^{\mathrm{new}}}{\mathrm{minimize}} \quad 
    & \left\| \boldsymbol{v}_i^{\mathrm{new}} - \hat{\boldsymbol{v}}_i(\boldsymbol{p}_i) \right\|_2^2 \\
    \mathrm{subject\ to} \quad 
    & \boldsymbol{H}_{ij} \boldsymbol{v}_i^{\mathrm{new}} \leq \boldsymbol{k}_{ij}, \quad j \in \mathcal{N}_i ,
\end{aligned}
}
\end{equation}
where $\mathcal{N}_i = [\mathbb{N}] \setminus \{i\}$ denotes the set of all robots except robot~$i$.
For ORCA, each neighbor $j$ contributes a single linear constraint, given by
\begin{equation}
\boldsymbol{H}_{ij} = \boldsymbol{a}_{ij}^\top \in \mathbb{R}^{1 \times 2}, 
\quad 
\boldsymbol{k}_{ij} = b_{ij} \in \mathbb{R},
\end{equation}
where $\boldsymbol{a}_{ij} := -\boldsymbol{n}$ and 
$b_{ij} := -\boldsymbol{n}^\top \boldsymbol{v}_i^{\mathrm{opt}} - \frac{\|\boldsymbol{u}\|_2}{2}$, 
as derived from Eq. \eqref{equa:min_u_orca}.
For LRCA, two linear constraints are imposed for each neighbor, leading to
\begin{equation}
\boldsymbol{H}_{ij} =
\begin{bmatrix}
    \boldsymbol{a}_{ij}^\top \\[3pt]
    \boldsymbol{c}_{ij}^\top
\end{bmatrix}
\in \mathbb{R}^{2 \times 2},
\quad
\boldsymbol{k}_{ij} =
\begin{bmatrix}
    b_{ij} \\[3pt]
    d_{ij}
\end{bmatrix}
\in \mathbb{R}^{2},
\end{equation}
where $\boldsymbol{c}_{ij} := \boldsymbol{h}^\top$ and 
$d_{ij} := \boldsymbol{h}^\top \boldsymbol{v}_i^{\mathrm{opt}}$ 
follow from Eq. \eqref{equa:rela_v}.
To circumvent Eq. \eqref{equa:qp}, we use duality theory, specifically KKT conditions of to derive an explicit expression for ${\boldsymbol{v}_i^{new}}^*$ as a function of the robots’ positions. These conditions are necessary and sufficient for a global optimum of this QP. The Lagrangian is:


\begin{equation}
\label{eq:lagrangian}
\mathcal{L}(\boldsymbol{v}_i^{\mathrm{new}}, \boldsymbol{\Lambda}) = \displaystyle
\left\| \boldsymbol{v}_i^{\mathrm{new}} - \hat{\boldsymbol{v}}_i \right\|^2 
+ \sum_{j \in \mathcal{N}_i} \boldsymbol{\lambda}_{ij}^\top\left(\boldsymbol{H}_{ij}\boldsymbol{v}_i^{\mathrm{new}} - \boldsymbol{k}_{ij}\right),
\end{equation}
where $\mathcal{N}_i = [\mathbb{N}] \setminus \{i\}$, and $\boldsymbol{\lambda}$ denotes the Lagrange multipliers. For ORCA $\boldsymbol{\lambda}_{ij}=\lambda_{ij}^{(a)}\in\mathbb{R}$, and for LRCA $\boldsymbol{\lambda}_{ij}=\begin{bmatrix}
    \lambda_{ij}^{(a)} \\[3pt]
    \lambda_{ij}^{(c)}
\end{bmatrix}
\in \mathbb{R}^{2}$. Let $\left( \boldsymbol{v}_i^{\mathrm{new}*}, \boldsymbol{\Lambda}^* \right)$ be the optimal primal-dual solutions to Eq. \eqref{equa:qp}, 

the KKT conditions are:

1. \textbf{Stationarity:} 
   \[
   \left.\nabla_{\boldsymbol{v}_i^{\mathrm{new}}} \mathcal{L}(\boldsymbol{v}_i^{\mathrm{new}}, \boldsymbol{\Lambda})\right|_{(\boldsymbol{v}_i^{\mathrm{new}*})} = \boldsymbol{0}
   \]
   \begin{equation}
   \label{eq:stationarity}
   \Longrightarrow \quad 
   \boldsymbol{v}_i^{\mathrm{new}*} =\displaystyle
       \hat{\boldsymbol{v}}_i - \frac{1}{2} \sum_{j \in \mathcal{N}_i} \boldsymbol{\lambda}_{ij}^{\top} \boldsymbol{H}_{ij}.
   \end{equation}

2. \textbf{Primal Feasibility:}
   \begin{equation}
   \boldsymbol{H}_{ij}^\top \boldsymbol{v}_i^{\mathrm{new}*} \leq \boldsymbol{k}_{ij},\quad\forall j \in \mathcal{N}_i.
   \end{equation}

3. \textbf{Dual Feasibility:}
   \begin{equation}
   \boldsymbol{\lambda}_{ij}^* \geq 0,\quad\forall j \in \mathcal{N}_i.
   \end{equation}

4. \textbf{Complementary Slackness:}
   \begin{equation}
   \boldsymbol{\lambda}_{ij}^* \left( \boldsymbol{H}_{ij} \boldsymbol{v}_i^{\mathrm{new}*} - \boldsymbol{k}_{ij} \right) = \boldsymbol{0},\quad\forall j \in \mathcal{N}_i.
   \end{equation}

\begin{definition}
Robot i is in deadlock if ${\boldsymbol{v}_i}^* = 0$ and the  prescribed nominal control $\hat{\boldsymbol{v}}_i \neq \mathbf{0} \Longleftrightarrow \boldsymbol{p}_i \neq \boldsymbol{p}_{d_i}$ and the speed remains at 0.
\end{definition}

A robot is in deadlock if it receives zero velocity commands from the QP controller indefinitely without having reached its goal. Importantly, deadlock is a system-level phenomenon and cannot occur for a single robot in isolation.

\subsection{Deadlock Conditions in Two-Robot ORCA Interactions}

According to the deadlock definition, establishing deadlock in ORCA reduces to identifying a two-robot configuration in which, despite nonzero target velocities, ORCA outputs zero velocities at two consecutive time steps. Since ORCA is memoryless and depends only on the current state, zero velocity at two consecutive steps implies that the robot velocity converges to zero thereafter.
Assume that a deadlock has occurred. Then there exist two consecutive time steps $t_1$ and $t_2$ such that
\[
\boldsymbol{v}_i^{t_1}=\boldsymbol{v}_i^{t_2}=\mathbf{0},
\qquad
\hat{\boldsymbol{v}}_i^{opt}=\hat{\boldsymbol{v}}_i^{t_1}\neq\mathbf{0}.
\]

The equality of the target velocities follows from the fact that the target velocity depends only on the robot’s current position and goal position; since $\boldsymbol{v}_i^{t_1}=0$, the robot position does not change between $t_1$ and $t_2$.

We consider a two-robot interaction in which robot $i$ is constrained only by robot $j$. Hence, at the optimum of the ORCA quadratic program, at most one collision-avoidance constraint is active. From the stationarity condition of the KKT optimality conditions, the ORCA solution satisfies
\[
\boldsymbol{v}_i^{*}
=
\hat{\boldsymbol{v}}_i
-\frac{1}{2}\lambda_{ij}^{*}\boldsymbol{a}_{ij}.
\]

Substituting $\boldsymbol{v}_i^{t_1}=\mathbf{0}$ yields
\begin{equation}
\label{eq:orca_dead_1_re}
\hat{\boldsymbol{v}}_i^{opt}
=
\frac{1}{2}\lambda_{ij}^{opt*}\boldsymbol{a}_{ij}^{opt},
\end{equation}
and by the same reasoning at time $t_2$,
\begin{equation}
\label{eq:orca_dead_2_re}
\hat{\boldsymbol{v}}_i^{t_1}
=
\frac{1}{2}\lambda_{ij}^{t_1*}\boldsymbol{a}_{ij}^{t_1}.
\end{equation}

When the relative velocity satisfies $\boldsymbol{v}_{i\mid j}^{t_1}=\mathbf{0}$, the ORCA formulation implies that the normal vector $\boldsymbol{n}_{ij}^{t_1}$ is aligned with the relative position vector,
\[
\exists\,\alpha>0 \ \text{s.t.}\ \boldsymbol{n}_{ij}^{t_1}=\alpha(\boldsymbol{p}_i-\boldsymbol{p}_j).
\]

Since $\boldsymbol{a}_{ij}^{t_1}=-\boldsymbol{n}_{ij}^{t_1}$ and $\lambda_{ij}^{t_1*}\ge0$, Eq.~\eqref{eq:orca_dead_2_re} implies that the target velocity $\hat{\boldsymbol{v}}_i^{t_1}$ is aligned with relative position vector $\boldsymbol{p}_j-\boldsymbol{p}_i$. Since $\hat{\boldsymbol{v}}_i^{t_1} = \hat{\boldsymbol{v}}_i^{opt}$, combining the above result with Eq.~\eqref{eq:orca_dead_1_re} implies that the constraint normal $\boldsymbol{n}_{ij}^{opt}$ is also aligned with the relative position vector $\boldsymbol{p}_j - \boldsymbol{p}_i$. According to the ORCA construction, alignment of the constraint normal with the relative position further implies that the relative velocity $\boldsymbol{v}_{i\mid j}^{opt}$ is colinear with the same direction. Hence
\[
\exists\,\alpha>0 \ \text{s.t.}\ 
\boldsymbol{v}_{i\mid j}^{opt}
=
\alpha(\boldsymbol{p}_j-\boldsymbol{p}_i).
\]

\begin{figure}[!t]
    \centering
    \includegraphics[width=1\linewidth]{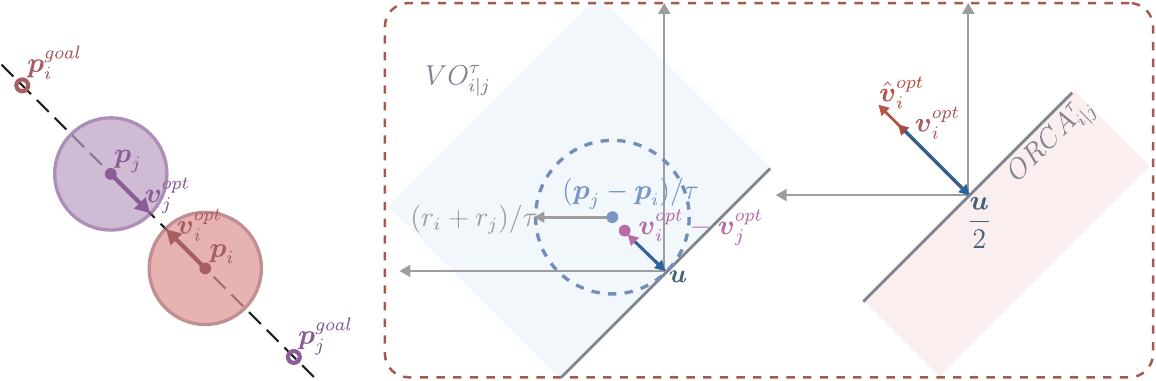}
    \caption{An example of deadlock in multi-robot navigation with two robots using the ORCA algorithm, along with the ORCA algorithm's computation process.}
    \label{fig:deadlock}
\end{figure}
The resulting geometric configuration is illustrated in Fig.~\ref{fig:deadlock}. Deadlock under ORCA occurs when the relative velocity and relative position of the two robots are parallel, while the target velocities of both robots point toward each other.

\subsection{Proving Deadlock-Free Behavior in Two-Robot LRCA}
\label{sec:prov_lrca}
\begin{theorem}
    Under the constraints of the random obstacle avoidance strategy of LRCA, the robot will not encounter deadlocks. 
\end{theorem}

\begin{proof}
    Consider a pair of robots $i$ and $j$. At a certain moment, if the velocities $\boldsymbol{v}_i^{new}$ and $\boldsymbol{v}_j^{new}$ calculated by the random obstacle avoidance strategy are both 0, then next time step, their relative velocity $\boldsymbol{v}_{i\mid j}^{opt}$at that moment is also 0 and $(\boldsymbol{v}_{i\mid j}^{opt}\times\delta\boldsymbol{v}_{i})\cdot(\boldsymbol{v}_{i\mid j}^{opt}\times\delta\boldsymbol{v}_{j}) = 0$. According to the second type of constraint, the constraint on robot i as:

    \begin{equation}
        \label{equa：}\boldsymbol{n}^\top\boldsymbol{v}_i^{new}\ge 0.
    \end{equation}
    
    According to Lagrangian and KKT conditions, $\boldsymbol{v}_i^{new}$ can be calculated:

    \begin{equation}
    {\boldsymbol{v}_i^{new}}^*=\hat{\boldsymbol{v}}_i+\frac{1}{2} \lambda_{i j}^* \boldsymbol{n}.
    \end{equation}
    
    
    

    
    
    
    
    
    According to Eq. \eqref{alg:probability}, there is a one-half probability that $\boldsymbol{n} = \mathbf{R}\left( \frac{\pi}{2} + \theta \right) \boldsymbol{e}_{ij}$, and a one-half probability that $\boldsymbol{n} = \mathbf{R}\left( -\frac{\pi}{2} - \theta \right) \boldsymbol{e}_{ij}$. if $\boldsymbol{n} = \mathbf{R}\left( \frac{\pi}{2} + \theta \right) \boldsymbol{e}_{ij}$, then $\boldsymbol{v}_i^{new}=0$, then when $\boldsymbol{n} = \mathbf{R}\left( -\frac{\pi}{2} - \theta \right) \boldsymbol{e}_{ij}$, $\boldsymbol{v}_i^{new}$ definitely is not 0. $\boldsymbol{v}_i^{new}$ will be zero in both cases if and only if $\hat{\boldsymbol{v}}_i=0$. Therefore, under the random obstacle avoidance algorithm, the robot will not get stuck.




\end{proof}

\section{Simulation Experiments}
This section details the experimental setup, showcases representative deadlock scenarios to validate LRCA, and evaluates its performance through both quantitative metrics and qualitative case studies across diverse navigation scenarios.

\subsection{Experiments Setup}

Implemented as an ORCA extension within the RVO library\footnote{\url{https://github.com/sybrenstuvel/Python-RVO2}}, LRCA sets a maximum robot speed of $1\mathrm{m/s}$ and a control step of $0.1\mathrm{s}$. Two prediction horizons are used: a short horizon ($0.3\mathrm{s}$) for ORCA-based avoidance and a long horizon ($1.5\mathrm{s}$) for LRCA. Robots' radius are uniformly sampled from $[0.1, 0.15]\,\mathrm{m}$ m.
Experiments cover three benchmark scenarios: Cross, Random, and Corridor. In the Cross Scenario, robots move along circle diameters and cross at the center, creating a highly challenging navigation environment. In the Random Scenario, start and goal positions are uniformly sampled in a rectangle. In the Corridor Scenario, robots traverse a narrow passage from opposite ends, simulating constrained navigation.
Performance is assessed by success rate, extra distance, extra time, and average speed. Success rate measures robots reaching goals collision-free within a time limit. Extra distance and time measure deviations from the ideal shortest-path distance and minimal travel time, respectively. Average speed is the mean navigation speed.
LRCA is compared against four representative circular-robot collision avoidance methods: multi-robot algorithms ORCA~\cite{b10} and RL-RVO~\cite{b22}, and social navigation algorithms T-MPC~\cite{b24,b25} and SARL~\cite{b23}. Although originally designed for ORCA-simulated pedestrians, the latter two are reasonably adapted to multi-robot settings and thus included for comparison.

\subsection{Analysis of Deadlock Cases under Cross Scenarios}

\begin{figure}[!t]
    \centering
    \includegraphics[width=1\linewidth]{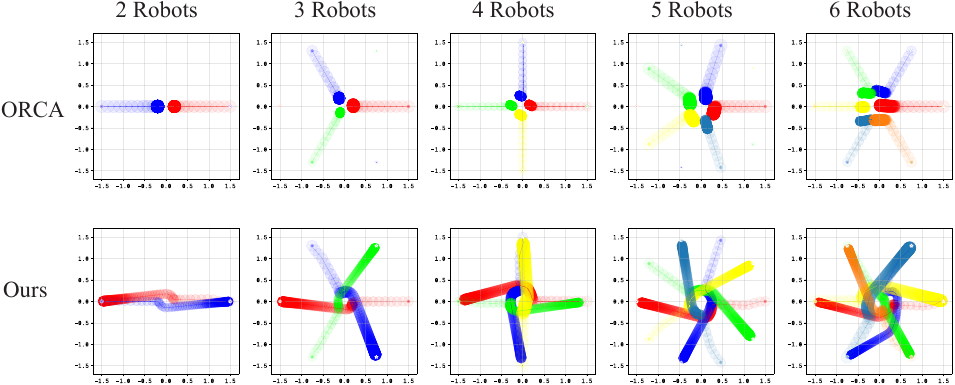}
    \caption{Comparison of the trajectories of ORCA and Ours in five scenarios prone to deadlock.}
    \label{fig:deadlock2}
\end{figure}

In center-symmetrical cross scenarios, the ORCA algorithm is prone to deadlock, as illustrated in Fig.~\ref{fig:deadlock2}. In this symmetric configuration, each robot moves along the straight line connecting its start and goal positions. To avoid collision, ORCA enforces symmetric deceleration, causing both robots to gradually slow down and eventually stop at the safety boundary with zero relative velocity. As a result, the robots fail to reach their goals despite having nonzero target velocities.
In contrast, the proposed LRCA strategy resolves this issue by explicitly breaking symmetry through randomized lateral avoidance. As shown in Fig.~\ref{fig:deadlock2}, robots under LRCA perform smooth lateral maneuvers and pass each other without stopping. By restricting velocity adjustments to a randomly selected side of the relative velocity, LRCA induces early asymmetric behavior, thereby preventing convergence to the zero-velocity equilibrium responsible for deadlock.
These results indicate that in highly symmetric cross scenarios, obstacle avoidance strategies based on minimal deviation, such as ORCA, are inherently susceptible to system deadlock. Incorporating a social-norm-inspired lateral avoidance mechanism, as in LRCA, effectively eliminates this failure mode.

\subsection{Quantitative Analysis of Navigation Performance under Cross Scenarios}

\begin{table}[t]
\caption{Performance metrics (as mean) evaluated for different methods on the cross scenarios with varied scene sizes and different number of robots.}
\resizebox{\textwidth}{!}{%
\begin{tabular}{@{}llllllllll@{}}
\toprule
\multirow{2}{*}{Metrics}                          & \multirow{2}{*}{Method} & \multicolumn{8}{l}{\#agents (radius of the scene)} \\ \cmidrule(l){3-10} 
                                                  &                         & 4 (2.0 m)        & 6 (2.0 m)       & 8 (2.5 m)      & 10 (2.5 m)       & 20 (3.0 m)       & 30 (4.0 m)      & 40 (5.0 m)      & 50 (6.0 m)      \\ \midrule
\multicolumn{1}{c}{\multirow{5}{*}{Success Rate(\%)}} & ORCA                    &\textbf{100.00}      &\textbf{100.00}      &\textbf{100.00}      &\textbf{100.00}      &79.35      &77.27      &67.33      &70.30      \\
\multicolumn{1}{c}{}                              & T-MPC                   &43.00      &16.67      &18.09     &10.00      &7.50     &-      &-      &-      \\
\multicolumn{1}{c}{}                              & SARL                    &\textbf{100.00}      &\textbf{100.00}      &\textbf{100.00}      &99.80      &94.00      &30.63      &69.50      &36.40      \\
\multicolumn{1}{c}{}                              & RL-RVO                  &\textbf{100.00}      &99.83      &96.50      &99.70      &95.85      &93.97     &91.18      &86.67      \\
\multicolumn{1}{c}{}                              & Ours                    &\textbf{100.00}      &\textbf{100.00}      &\textbf{100.00}      &99.60      &\textbf{97.80}      &\textbf{97.00}      &\textbf{95.05}      &\textbf{95.30}      \\ \midrule
\multirow{5}{*}{Extra Distance(m)}                         & ORCA                    &4.35      &0.40      &0.66     &1.09      &1.25      &1.46      &1.77     &2.83      \\
                                                  & T-MPC                   &5.08      &4.34      &5.07      &5.06      &4.88      &-      &-      &-      \\
                                                  & SARL                    &15.63      &15.77      &13.90      &13.70     &19.95      &19.81      &6.29      &18.25     \\
                                                  & RL-RVO                  &1.59      &1.62      &1.74      &1.97      &3.15      &3.60      &4.04      &4.57      \\
                                                  & Ours                    &\textbf{0.26}      &\textbf{0.30}      &\textbf{0.36}      &\textbf{0.45}      &\textbf{0.74}      &\textbf{1.05}      &\textbf{1.35}      &\textbf{1.37}      \\ \midrule
\multirow{5}{*}{Extra Time(s)}                             & ORCA                    &78.00     &8.30      &6.41      &6.83      &5.45      &6.91      &8.07      &12.04      \\
                                                  & T-MPC                   &4.98      &4.24      &4.97      &4.96      &4.48      &-      &-      &-      \\
                                                  & SARL                    &18.45      &18.54      &18.39      &18.56     &26.98      &25.35      &10.22      &26.03      \\
                                                  & RL-RVO                  &6.26      &5.80      &6.33      &7.23      &12.00      &14.04      &15.99      &19.22      \\
                                                  & Ours                    &\textbf{2.91}      &\textbf{3.19}      &\textbf{3.36}      &\textbf{3.59}      &\textbf{4.38}      &\textbf{5.32}      &\textbf{6.20}      &\textbf{6.27}      \\ \midrule
\multirow{5}{*}{Average Speed(m/s)}                    & ORCA                    &0.18     &0.59      &0.65      &0.56      &0.62     &0.62      &0.56      &0.49      \\
                                                  & T-MPC                   &0.59      &0.61      &0.63      &0.59      &0.70      &-      &-      &-      \\
                                                  & SARL                    &\textbf{0.85}      &\textbf{0.86}      &\textbf{0.83}      &\textbf{0.80}      &\textbf{0.79}      &\textbf{0.83}      &\textbf{0.80}      &\textbf{0.79}      \\
                                                  & RL-RVO                  &0.37      &0.38      &0.27      &0.29      &0.17      &0.20      &0.24      &0.30  \\
                                                  & Ours                    &0.80      &0.78      &0.75      &0.73      &0.66      &0.62      &0.58      &0.58      \\ \bottomrule
\end{tabular}%
}
\end{table}

In the cross scenario, LRCA consistently outperforms other methods across all metrics. The success rate for LRCA nearly 100\% for all group sizes, while ORCA sees a significant decline, dropping from 100\% to 70.30\% as the number of robots increases. T-MPC and SARL also struggle in larger groups, with SARL’s success rate decreasing to 36.4\% for 50 robots, and T-MPC's success rate dropping to 0 when the number of robots exceeds 30.
For extra distance, LRCA shows the best performance, with values ranging from 0.26 meters for 4 robots to 1.37 meters for 50 robots. ORCA and T-MPC have higher extra distances, especially with more robots.
Regarding extra time, LRCA again leads, with values ranging from 2.91 seconds for 4 robots to 6.27 seconds for 50 robots. ORCA and RL-RVO experience higher extra times as the robot count increases, while SARL reaches the highest extra time of 26.98 seconds for 20 robots.

Finally, in terms of average speed, LRCA maintains high speeds, from 0.8 m/s for 4 robots to 0.58 m/s for 50 robots. In contrast, ORCA exhibits significantly lower navigation speeds across all scenarios due to the frequent occurrence of deadlock. RL-RVO shows unstable navigation speeds, which can be attributed to the limited interpretability of its learned policy. Although SARL achieves the highest navigation speed in all scenarios, this excessive speed leads to reduced navigation safety and efficiency when other performance metrics are considered.
Overall, LRCA demonstrates superior performance in cross scenarios, consistently maintaining high success rates, low distance and time deviations, and robust navigation behavior even as the number of robots increases.

\subsection{Qualitative Analysis of Navigation Trajectories}
\begin{figure}[!t]
    \centering
    \includegraphics[width=1\linewidth]{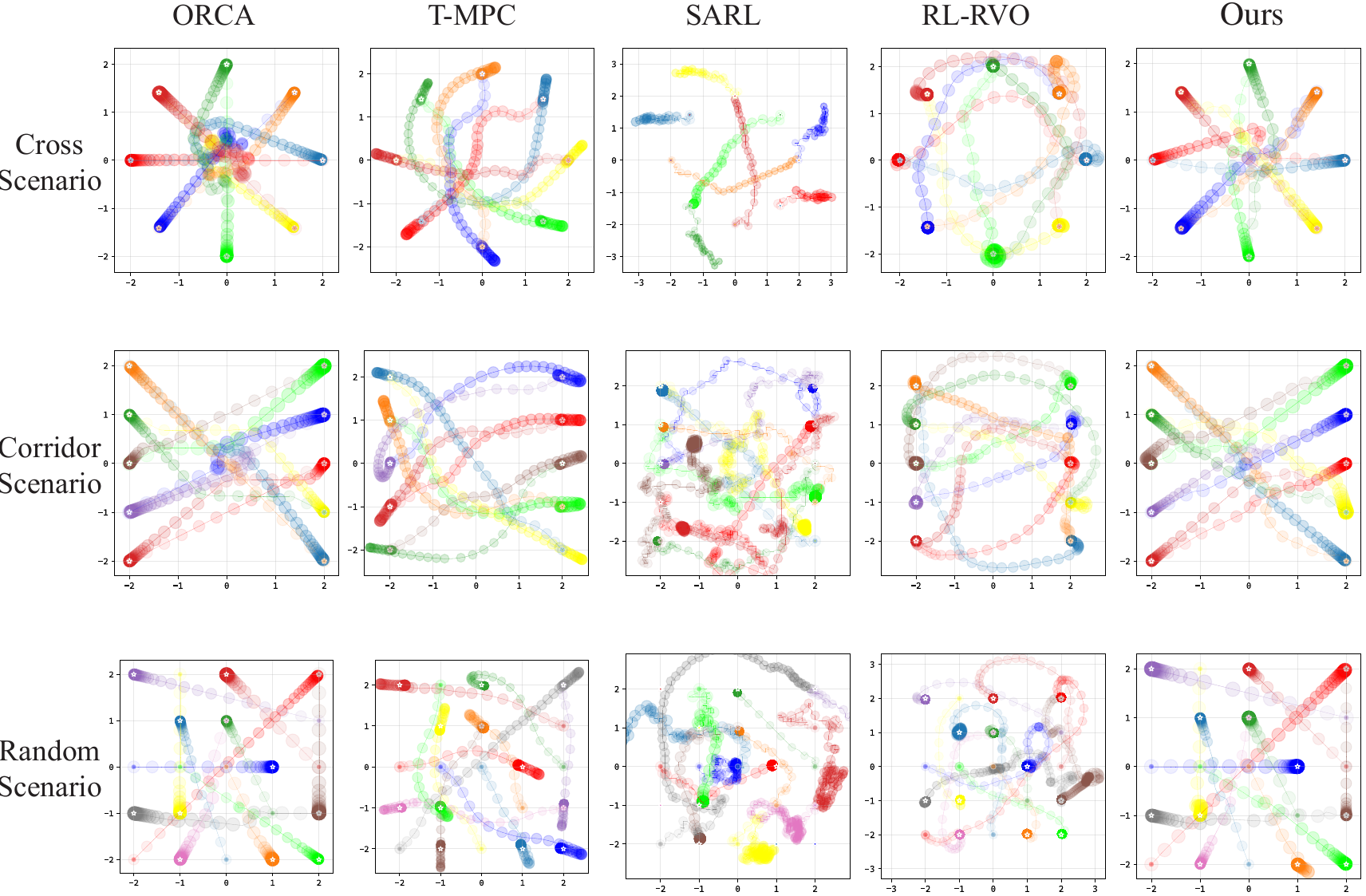}
    \caption{Navigation trajectories of five methods (ORCA, T-MPC, SARL, RL-RVO, Ours) in cross scenario, corridor scenario, and random scenario.}
    \label{fig:traj}
\end{figure}

Fig. \ref{fig:traj} presents qualitative comparisons of navigation trajectories generated by ORCA, T-MPC, SARL, RL-RVO, and the proposed LRCA in three representative scenarios. Each colored curve denotes the trajectory of an individual robot from its start to its goal.
In the cross scenario (first row), ORCA exhibits strong symmetry and congestion near the intersection, leading to inefficient interactions. T-MPC alleviates conflicts but produces long and curved detours. SARL results in irregular and sometimes unstable trajectories, while RL-RVO tends to generate overly conservative circular motions. In contrast, LRCA produces smooth and well-structured trajectories by introducing consistent lateral avoidance, enabling robots to pass through the intersection efficiently.

In the corridor scenario (second row), ORCA frequently suffers from reciprocal blocking in the narrow passage. Learning-based methods reduce collisions but show oscillatory or overly cautious behaviors. LRCA maintains clear directional separation, allowing robots to traverse the corridor smoothly with minimal interference.
In the random scenario (third row), ORCA remains prone to mutual blocking under dense interactions, and SARL and RL-RVO generate scattered or unnecessarily long paths. LRCA achieves compact and well-separated trajectories, demonstrating robust collision avoidance without sacrificing efficiency.
Overall, the qualitative results indicate that LRCA effectively breaks symmetric interactions through lateral relative-velocity adjustments, leading to smoother, more efficient, and more reliable navigation across diverse scenarios.

\subsection{Computational Efficiency Analysis}

\begin{figure}[!t]
    \centering
    \includegraphics[width=1\linewidth]{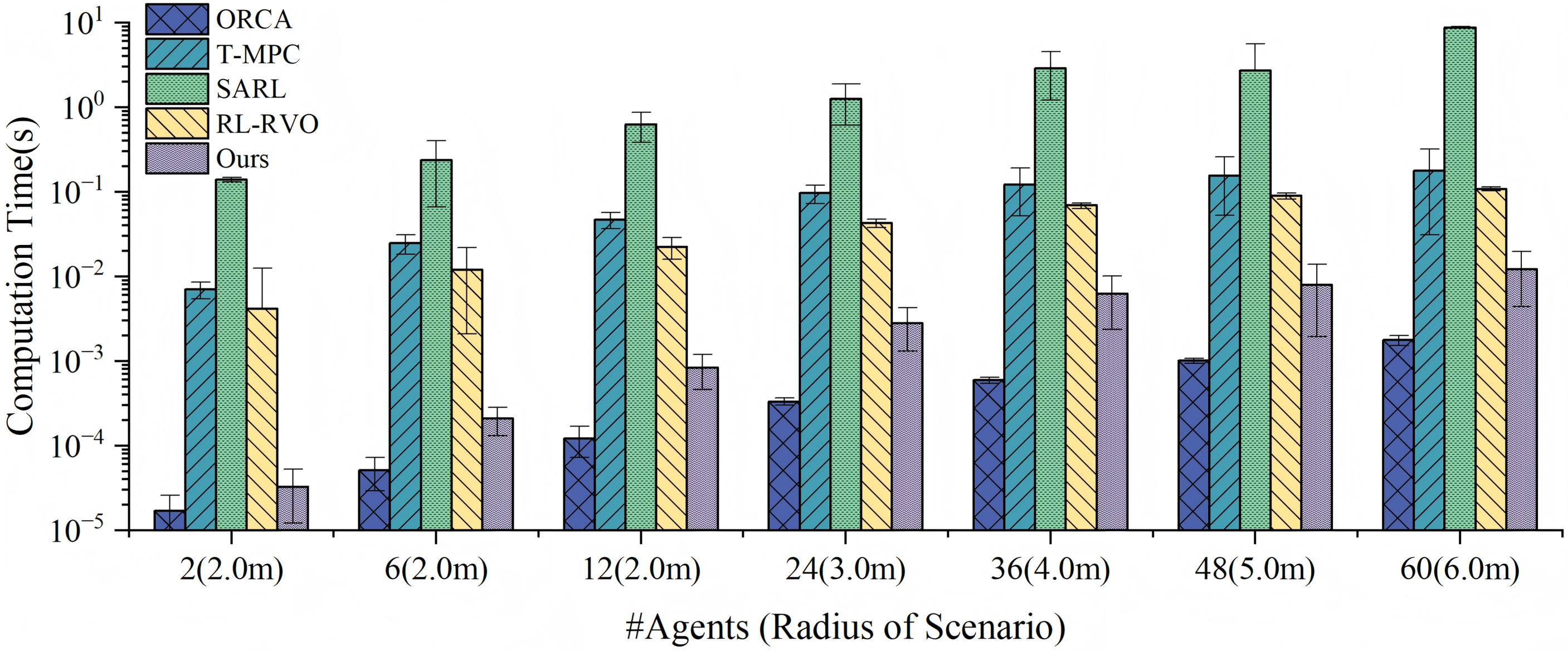}
    \caption{Computation time (as mean/SD in milliseconds) for different methods with varied number of agents.}
    \label{fig:time}
\end{figure}


Fig. \ref{fig:time} shows the computation time (mean/standard deviation in milliseconds) for different methods with varying numbers of agents. ORCA is the most computationally efficient method, with times increasing from  $1.69\times10^{-5}$ms for 2 agents to $1.76\times10^{-3}$ms for 60 agents. T-MPC has a higher computational cost, increasing from $7.01\times10^{-3}$ms to $1.55\times10^{-1}$ms as the number of agents grows. SARL exhibits the highest computational cost, rising from $1.39\times10^{-1}$ms for 2 agents to 8.75 seconds for 60 agents. RL-RVO has moderate computation times, ranging from $3.72\times10^{-3}$ms to $1.21\times10^{-1}$ms. Our method, LRCA, provides a good balance, with times increasing from $3.26\times10^{-5}$ms to $1.21\times10^{-2}$ms, making it efficient yet slightly slower than ORCA. Overall, LRCA substantially improves navigation safety and efficiency while preserving a level of computational lightness and real-time performance comparable to that of ORCA.

\section{Conclusions}

This paper proposed the Lateral Reciprocal Collision Avoidance (LRCA) algorithm for multi-robot navigation. Inspired by pedestrian behaviors, LRCA introduces lateral acceleration and randomized lateral velocity selection to achieve deadlock-free collision avoidance. By reformulating ORCA as a quadratic program and analyzing its KKT conditions, we identified deadlock scenarios in two-robot interactions and showed that LRCA effectively avoids such deadlocks.
Experimental results demonstrate that LRCA outperforms ORCA and other baseline methods in deadlock-prone scenarios, achieving improved navigability, efficiency, and safety across diverse navigation tasks.
A limitation of the current work is that the modeling and algorithm are restricted to two-dimensional space. Future work will extend LRCA to three-dimensional scenarios to enable deadlock-free reciprocal collision avoidance in 3D environments.

\begin{credits}
\subsubsection{\ackname} This work was supported by National Key R\&D Program of China (Grant No. 2025YFF0521600) and the National Natural Science Foundation of China (No. 52505001).

\end{credits}
%
%
%

\begin{thebibliography}{8}
\bibitem{Wang05052025}
Wang, G., Zhang, G., Zhou, X., Zhang, Y.: Distributed multi-robot task dynamic allocation in digital-twin factory towards industry 5.0. Int. J. Prod. Res. 1-24 (2025)
\bibitem{b2}
Choset, H., Lynch, K.M., Hutchinson, S., Kantor, G.A., Burgard, W.: Principles of robot motion: theory, algorithms, and implementations. MIT Press (2005)
\bibitem{b3}
Schwartz, J.T., Sharir, M.: On the piano movers' problem: III. Coordinating the motion of several independent bodies: The special case of circular bodies moving amidst polygonal barriers. Int. J. Robot. Res. 2(3), 46-75 (1983)
\bibitem{b4}
Yu, J., LaValle, S.M.: Optimal multirobot path planning on graphs: Complete algorithms and effective heuristics. IEEE Trans. Robot. 32(5), 1163-1177 (2016)
\bibitem{b5}
Tang, S., Thomas, J., Kumar, V.: Hold Or take Optimal Plan (HOOP): A quadratic programming approach to multirobot trajectory generation. Int. J. Robot. Res. 37(9), 1062-1084 (2018)
\bibitem{b6}
Bareiss, D., van den Berg, J.: Generalized reciprocal collision avoidance. Int. J. Robot. Res. 34(12), 1501-1514 (2015)
\bibitem{b7}
Claes, D., Hennes, D., Tuyls, K., Meeussen, W.: Collision avoidance under bounded localization uncertainty. In: IEEE/RSJ International Conference on Intelligent Robots and Systems, pp. 1192-1198. IEEE (2012)
\bibitem{b8}
Hennes, D., Claes, D., Meeussen, W., Tuyls, K.: Multi-robot collision avoidance with localization uncertainty. In: Proceedings of the International Conference on Autonomous Agents and Multiagent Systems, pp. 147-154 (2012)
\bibitem{b9}
Snape, J., van den Berg, J., Guy, S.J., Manocha, D.: The hybrid reciprocal velocity obstacle. IEEE Trans. Robot. 27(4), 696-706 (2011)
\bibitem{b10}
van den Berg, J., Guy, S.J., Lin, M., Manocha, D.: Reciprocal n-body collision avoidance. In: International Symposium on Robotics Research, pp. 3-19. Springer, Berlin (2011)
\bibitem{b11}
Fiorini, P., Shiller, Z.: Motion planning in dynamic environments using velocity obstacles. Int. J. Robot. Res. 17(7), 760-772 (1998)
\bibitem{b12}
van den Berg, J., Guy, S.J., Lin, M., Manocha, D.: Optimal reciprocal collision avoidance for multi-agent navigation. In: Proc. of the IEEE International Conference on Robotics and Automation, pp. 1-8. IEEE (2010)
\bibitem{b13}
Douthwaite, J.A., Zhao, S., Mihaylova, L.S.: A comparative study of velocity obstacle approaches for multiagent systems. In: 2018 UKACC 12th International Conference on Control (CONTROL), pp. 289-294. IEEE (2018)
\bibitem{b14}
Helbing, D., Molnar, P.: Social force model for pedestrian dynamics. Phys. Rev. E 51(5), 4282 (1995)
\bibitem{b15}
Moussaïd, M., Helbing, D., Theraulaz, G.: How simple rules determine pedestrian behavior and crowd disasters. Proc. Natl. Acad. Sci. USA 108(17), 6884-6888 (2011)
\bibitem{b16}
Chen, J., Sun, D.: Resource constrained multirobot task allocation based on leader-follower coalition methodology. Int. J. Robot. Res. 30(12), 1423-1434 (2011)
\bibitem{b17}
Sharon, G., Stern, R., Felner, A., Sturtevant, N.R.: Conflict-based search for optimal multi-agent pathfinding. Artif. Intell. 219, 40-66 (2015)
\bibitem{b18}
Li, J., Ruml, W., Koenig, S.: EECBS: A bounded-suboptimal search for multi-agent path finding. In: Proc. of the AAAI Conference on Artificial Intelligence, vol. 35, no. 14, pp. 12353-12362. AAAI Press (2021)
\bibitem{b19}
Okumura, K.: LaCAM: Search-based algorithm for quick multi-agent pathfinding. In: Proc. of the AAAI Conference on Artificial Intelligence, vol. 37, no. 10, pp. 11655-11662. AAAI Press (2023)
\bibitem{b20}
Okumura, K., Machida, M., Défago, X., Tamura, Y.: Priority inheritance with backtracking for iterative multi-agent path finding. Artif. Intell. 310, 103752 (2022)
\bibitem{b21}
Leitmann, G., Skowronski, J.: Avoidance control. J. Optim. Theory Appl. 23(4), 581-591 (1977)
\bibitem{b22}
Leitmann, G., Skowronski, J.: A note on avoidance control. Optim. Control Appl. Methods 4(4), 335-342 (1983)
\bibitem{b23}
Chen, C., Liu, Y., Kreiss, S., Alahi, A.: Crowd-robot interaction: Crowd-aware robot navigation with attention-based deep reinforcement learning. In: Proc. IEEE Int. Conf. on Robotics and Automation (ICRA), pp. 6015-6022. IEEE (2019)
\bibitem{b24}
Poddar, S., Mavrogiannis, C., Srinivasa, S.S.: From crowd motion prediction to robot navigation in crowds. In: 2023 IEEE/RSJ International Conference on Intelligent Robots and Systems (IROS), pp. 6765-6772. IEEE (2023)
\bibitem{b25}
Mavrogiannis, C., Balasubramanian, K., Poddar, S., Gandra, A., Srinivasa, S.S.: Winding through: Crowd navigation via topological invariance. IEEE Robot. Autom. Lett. 8(1), 121-128 (2023)
\bibitem{b26}
Stipanović, D.M., Hokayem, P.F., Spong, M.W., Šiljak, D.D.: Cooperative avoidance control for multiagent systems. J. Dyn. Syst. Meas. Control 129(5), 699-707 (2007)
\bibitem{b27}
Hokayem, P.F., Stipanović, D.M., Spong, M.W.: Coordination and collision avoidance for lagrangian systems with disturbances. Appl. Math. Comput. 217(3), 1085-1094 (2010)
\bibitem{b28}
van den Berg, J., Lin, M., Manocha, D.: Reciprocal velocity obstacles for real-time multi-agent navigation. In: Proc. IEEE Int. Conf. on Robotics and Automation (ICRA), pp. 1928-1935. IEEE (2008)
\bibitem{b29}
Duives, D.C., Daamen, W., Hoogendoorn, S.P.: State-of-the-art crowd motion simulation models. Transp. Res. Part C Emerg. Technol. 37, 193-209 (2013)
\bibitem{b30}
Schultz, M., Rößger, L., Fricke, H., Schlag, B.: Group dynamic behavior and psychometric profiles as substantial driver for pedestrian dynamics. In: Pedestrian and Evacuation Dynamics 2012, pp. 1097-1111. Springer, Cham (2013)
\bibitem{b31}
Grover, J.S., Liu, C., Sycara, K.: Deadlock analysis and resolution for multi-robot systems. In: International Workshop on the Algorithmic Foundations of Robotics (WAFR), pp. 294-312. Springer, Cham (2020)

\end{thebibliography}
%

\end{document}